\documentclass[10pt,twocolumn,letterpaper]{article}

\usepackage{cvpr}
\usepackage{graphicx}
\usepackage{times}
\usepackage{epsfig}
\usepackage{amsmath}
\usepackage{amssymb}
\usepackage{algorithm}
\usepackage{multirow}
\usepackage{mathtools,xparse}
\usepackage{siunitx}
\usepackage{stmaryrd}
\usepackage[noend]{algpseudocode}
\usepackage{placeins}
\usepackage{mdframed}
\usepackage{amsmath,amssymb} 
\usepackage{color}
\usepackage{subcaption}
\usepackage[table]{xcolor}

\newcommand{\trans}{\text{T}}
\newcommand*\rot{\rotatebox{90}}

\definecolor{g1}{rgb}{1.00,1.00,1.00}

\usepackage[pagebackref=true,breaklinks=true,letterpaper=true,colorlinks,bookmarks=false]{hyperref}

\cvprfinalcopy 

\def\cvprPaperID{4694} 

\begin{document}

\title{MAGSAC: Marginalizing Sample Consensus}

\author{Daniel Barath$^{12}$, Jiri Matas$^{1}$, and Jana Noskova$^{1}$\\
$^1$ Centre for Machine Perception, Department of Cybernetics \\
  Czech Technical University, Prague, Czech Republic \\
  $^2$ Machine Perception Research Laboratory, 
  MTA SZTAKI, Budapest, Hungary \\
    {\tt\small barath.daniel@sztaki.mta.hu}
}


\maketitle

\begin{abstract}
A method called, $\sigma$-consensus, is proposed to eliminate the need for a user-defined inlier-outlier threshold in RANSAC. Instead of estimating the noise $\sigma$, it is marginalized over a range of noise scales. 
The optimized model is obtained by weighted least-squares fitting where the weights come from the marginalization over $\sigma$ of the point likelihoods of being inliers. 
A new quality function is proposed not requiring $\sigma$ and, thus, a set of inliers to determine the model quality. Also, a new termination criterion for RANSAC is built on the proposed marginalization approach.
Applying $\sigma$-consensus, MAGSAC is proposed with no need for a user-defined $\sigma$ and improving the accuracy of robust estimation significantly.
It is superior to the state-of-the-art in terms of geometric accuracy on publicly available real-world datasets for epipolar geometry ($\textbf{F}$ and $\textbf{E}$) and homography estimation. 
In addition, applying $\sigma$-consensus only once as a post-processing step to the RANSAC output \textit{always improved} the model quality on a wide range of vision problems without noticeable deterioration in processing time, adding a few milliseconds.\footnote{The source code is at \url{https://github.com/danini/magsac}}
\end{abstract}

\section{Introduction}

The RANSAC (RANdom SAmple Consensus) algorithm proposed by Fischler and Bolles~\cite{fischler1981random} in 1981 has become the most widely used robust estimator in computer vision. RANSAC and its variants have been successfully applied to a wide range of vision tasks, e.g.\ motion segmentation~\cite{torr1993outlier}, short baseline stereo~\cite{torr1993outlier,torr1998robust}, wide baseline stereo matching~\cite{pritchett1998wide,matas2004robust,mishkin2015mods}, detection of geometric primitives~\cite{sminchisescu2005incremental}, image mosaicing~\cite{ghosh2016survey}, and to perform~\cite{zuliani2005multiransac} or initialize multi-model fitting~\cite{isack2012energy,pham2014interacting}.
In brief, the RANSAC approach repeatedly selects random subsets of the input point set and fits a model, e.g.\ a plane to three 3D points or a homography to four 2D point correspondences. Next, the quality of the estimated model is measured, for instance by the size of its support, i.e.\ the number of inliers. Finally, the model with the highest quality, polished e.g.\ by least squares fiting on its inliers, is returned.

Since the publication of RANSAC, a number of modifications has been proposed. 
NAPSAC~\cite{nasuto2002napsac}, PROSAC~\cite{chum2005matching} and EVSAC~\cite{fragoso2013evsac} modify the sampling strategy to increase the probability of selecting an all-inlier sample early. 
NAPSAC assumes that the inliers are spatially coherent, PROSAC exploits an a priori predicted inlier probability of the points and EVSAC estimates a confidence in each of them. 
MLESAC~\cite{torr2000mlesac} estimates the model quality by a maximum likelihood process with all its beneficial properties, albeit under certain assumptions about inlier and outlier distributions. 
In practice, MLESAC results are often superior to the inlier counting of plain RANSAC and they are less sensitive to the user-defined inlier-outlier threshold.
In MSAC~\cite{torr2002bayesian}, the robust estimation is formulated as a process that estimates both the parameters of the data distribution and the quality of the model in terms of maximum a posteriori. timates the model quality by a maximum likelihood process with all its beneficial properties, albeit under certain assumptions about inlier and outlier distributions. 


One of the highly attractive properties of RANSAC is its small number of control parameters.
The termination is controlled by a manually set confidence value $\eta$ and the sampling stops as soon as the probability of finding a model with higher support falls below $1-\eta$.\footnote{Note that the probabilistic interpretation of $\eta$ holds only for the standard $\{ 0, 1 \}$ cost function.} The setting of $\eta$ is not problematic, the typical values are 0.95 or 0.99, depending on the required confidence in the solution.

\begin{figure}
  	\centering
  	\begin{subfigure}[t]{0.49\columnwidth}
  	    \includegraphics[width=0.999\columnwidth]{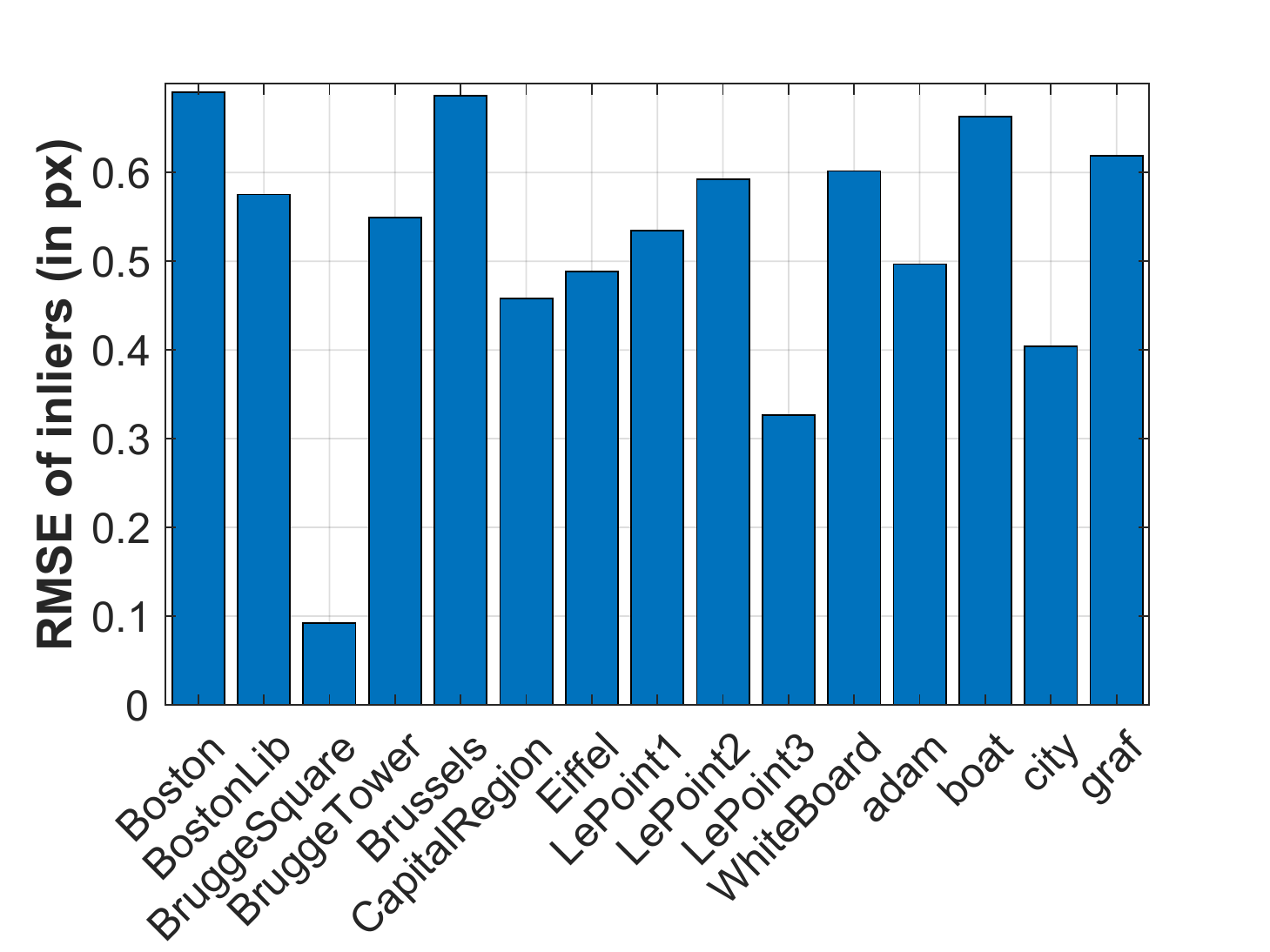}
  	    \caption{{\fontfamily{cmtt}\selectfont homogr} dataset}
    \end{subfigure}
  	\begin{subfigure}[t]{0.495\columnwidth}
  	    \includegraphics[width=0.999\columnwidth]{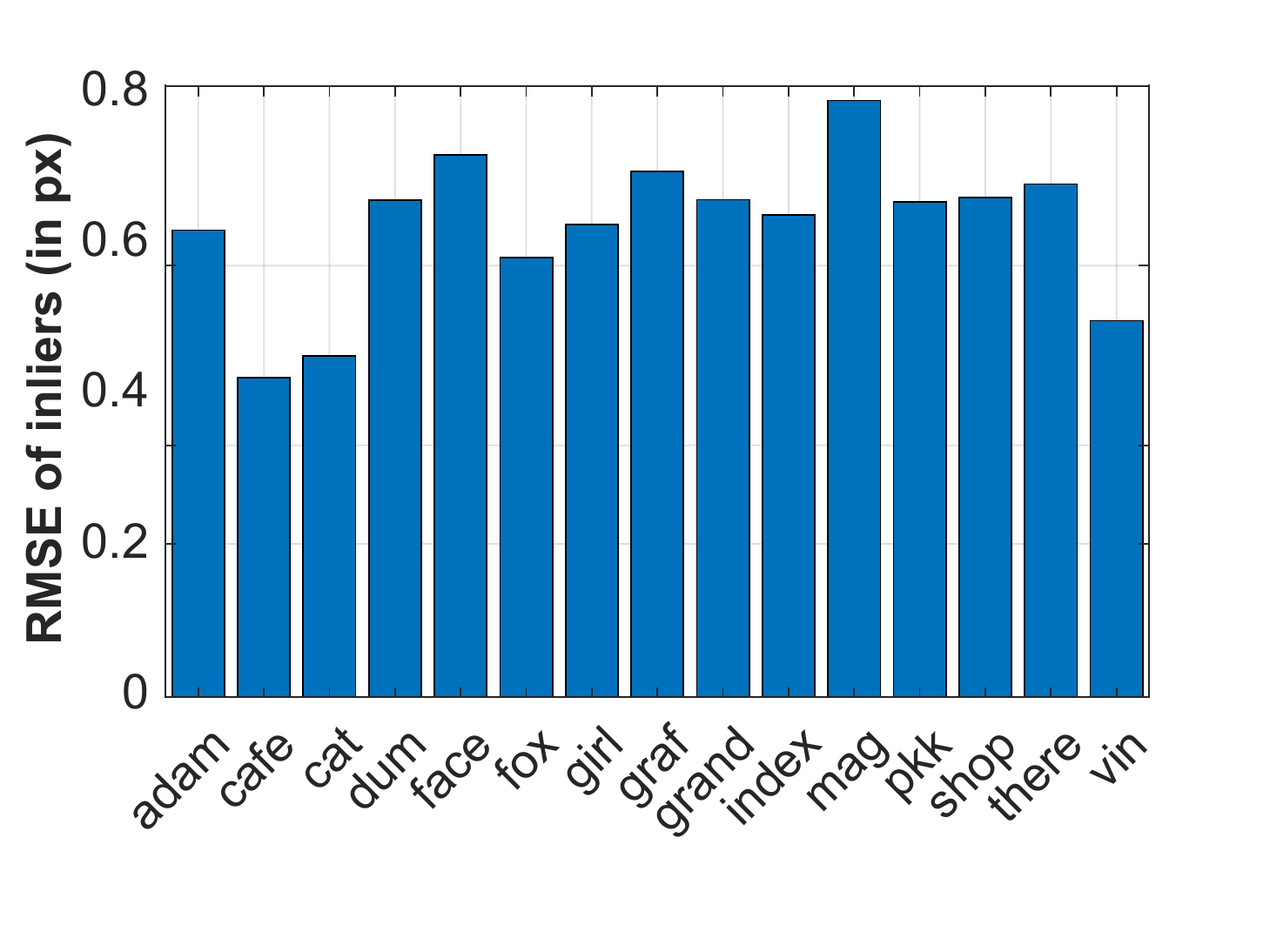}
  	    \caption{{\fontfamily{cmtt}\selectfont EVD} dataset}
    \end{subfigure}
  	\begin{subfigure}[t]{0.49\columnwidth}
  	    \includegraphics[width=0.999\columnwidth]{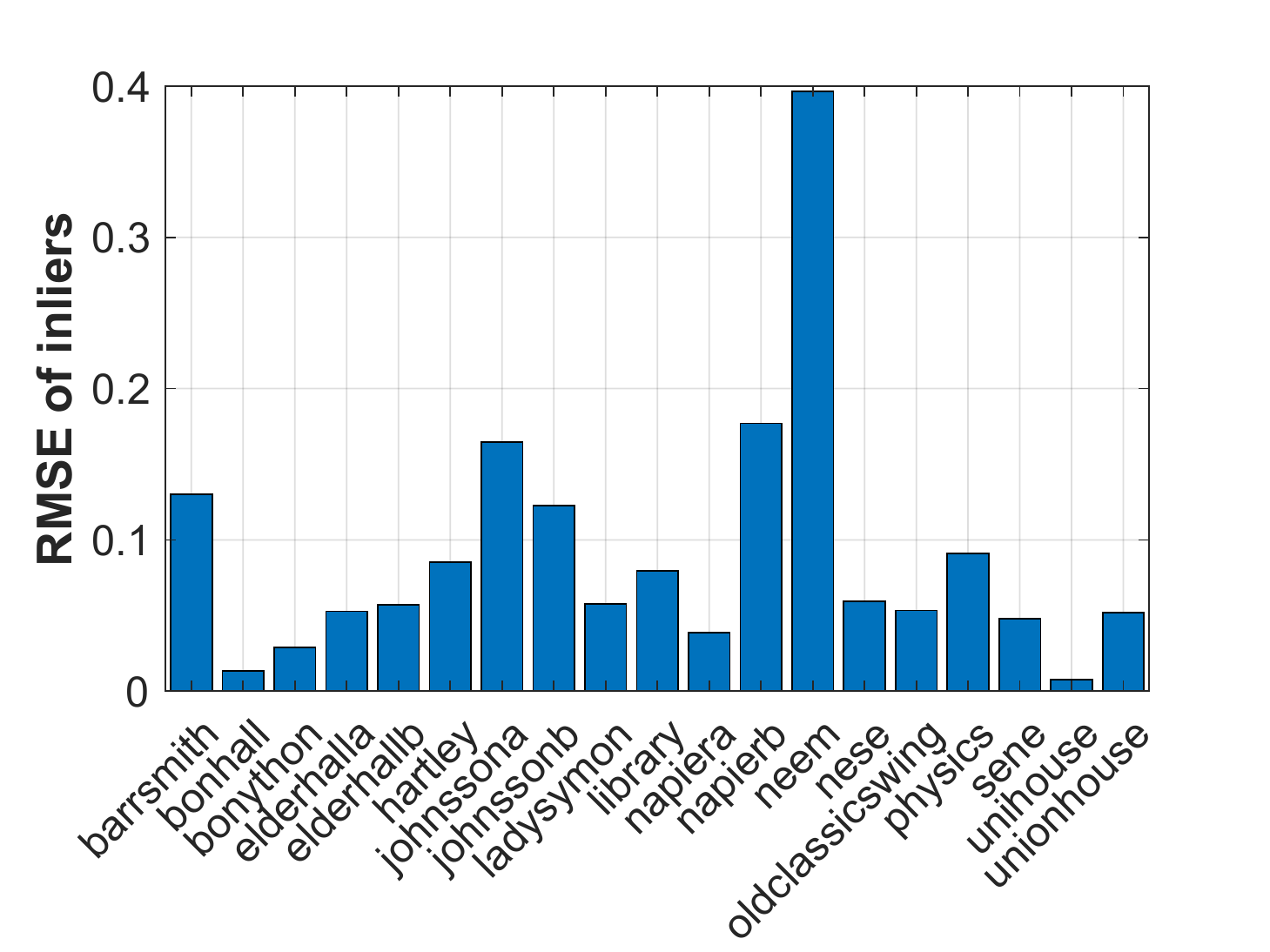}
  	    \caption{{\fontfamily{cmtt}\selectfont AdelaideRMF} dataset}
    \end{subfigure}
  	\begin{subfigure}[t]{0.49\columnwidth}
  	    \includegraphics[width=0.999\columnwidth]{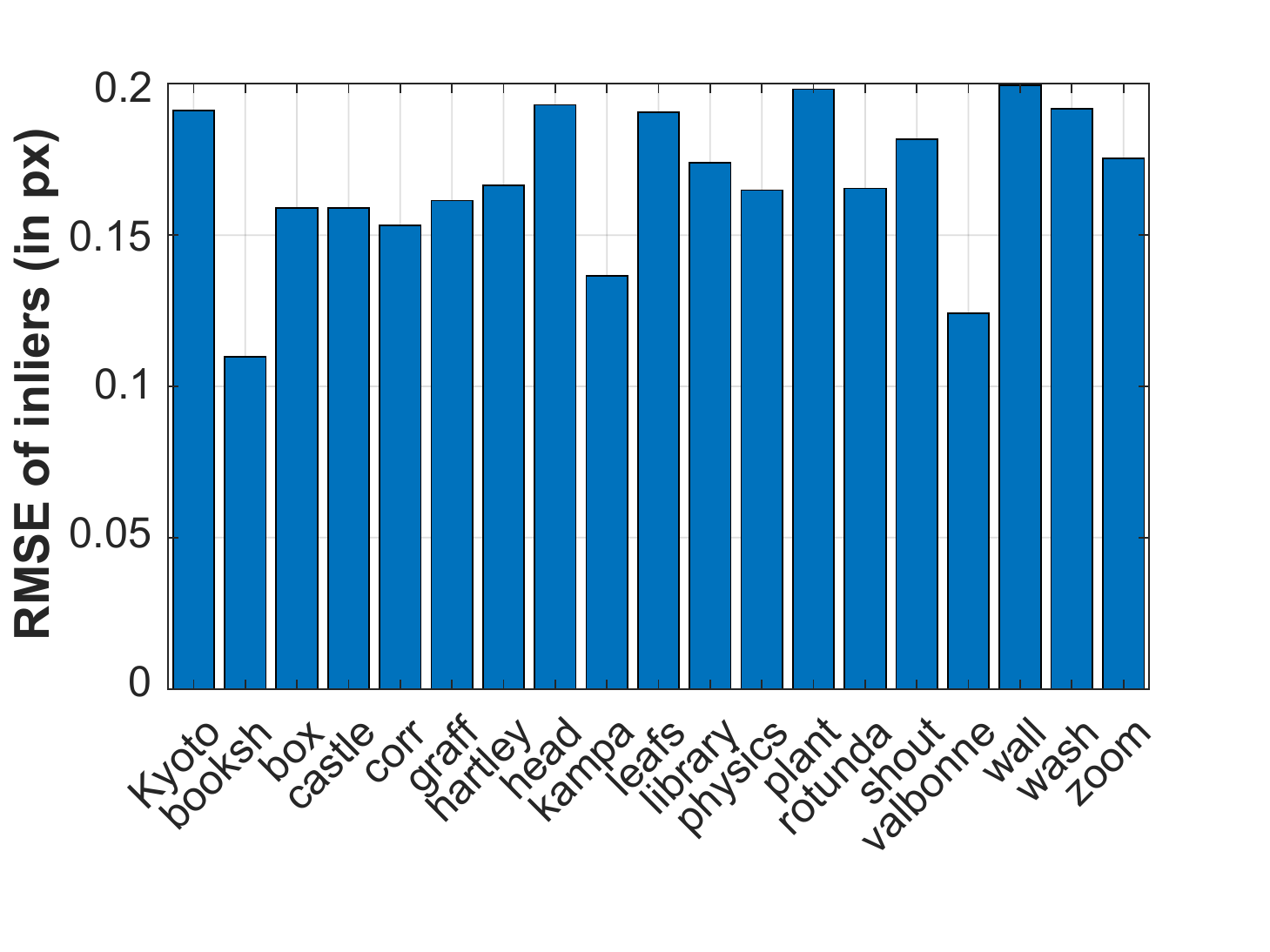}
  	    \caption{{\fontfamily{cmtt}\selectfont kusvod2} dataset}
    \end{subfigure}
    \caption{The average residuals (RMSE in pixels; vertical axis) of manually annotated inliers given the ground truth model for each scene (horizontal) of four datasets.}
    \label{fig:avg_residuals}
\end{figure}

The second, and most critical, parameter is the inlier noise scale $\sigma$ that determines the inlier-outlier threshold $\tau(\sigma)$ which strongly influences the outcome of the procedure. 
In standard RANSAC and its variants, $\sigma$ must be provided by the user which limits its fully automatic out-of-the-box use and requires the user to acquire knowledge about the problem at hand.
In Fig.~\ref{fig:avg_residuals}, the inlier residuals are shown for four real datasets demonstrating that $\sigma$ varies scene-by-scene and, thus, there is no single setting which can be used for all cases.   

To reduce the dependency on this threshold, MINPRAN~\cite{stewart1995minpran} assumes that the outliers are uniformly distributed and finds the model where the inliers are least likely to have occurred randomly. Moisan et al.~\cite{moisan2012automatic} proposed a contrario RANSAC, to optimize each model by selecting the most likely noise scale.

As the \textit{major contribution} of this paper, we propose an approach, $\sigma$-consensus, that eliminates the need for $\sigma$, the noise scale parameter. Instead of $\sigma$, only an upper limit is required. 
The final outcome is obtained by weighted least-squares fitting, where the weights are given for {\it marginalizing} over $\sigma$, using likelihood of the model given data and $\sigma$. Besides finessing the need for a precise scale parameter, the novel method, called MAGSAC, is more precise than previously published RANSACs.
Also, we propose a post-processing step applying $\sigma$-consensus to the \textit{so-far-the-best-model} without noticeable deterioration in processing time, i.e.\ at most a few milliseconds. 
In our experiments, the method \textit{always improved} the input model (coming from RANSAC, MSAC or LO-RANSAC) on a wide range of problems. Thus we see no reason for not applying it after the robust estimation finished.
As a \textit{second contribution}, we define a new quality function for RANSAC. It measures the quality of a model without requiring $\sigma$ and, therefore, a set of inliers to measure the model quality.
Moreover, as a \textit{third contribution}, due to not having a single inlier set and, thus, an inlier ratio, the standard termination criterion of RANSAC is marginalized over $\sigma$ to be applicable to the proposed method.

\section{Notation}
 
In this paper, the input points are denoted as $\mathcal{P} = \{ p \; | \; p \in \mathbb{R}^k, k \in \mathbb{N}_{>0} \}$, where $k$ is the dimension, e.g.\ $k=2$ for 2D points and $k = 4$ for point correspondences. The inlier set is $\mathcal{I} \subseteq \mathcal{P}$. 
The model to fit is represented by its parameter vector $\theta \in \Theta$, where $\Theta = \{ \theta \; | \; \theta \in \mathbb{R}^d, d \in \mathbb{N}_{>0} \}$ is the manifold, for instance, of all possible 2D lines and $d$ is dimension of the model, e.g.\ $d = 2$ for 2D lines (angle and offset). 
Fitting function $F : \mathcal{P}^* \to \Theta$ calculates the model parameters from $n \geq m$ points, where $\mathcal{P}^* = \exp \mathcal{P}$ is the power set of $\mathcal{P}$ and $m \in \mathbb{N}_{>0}$ is the minimum point number for fitting a model, e.g.\ $m = 2$ for 2D lines. Note that $F$ is a combined function applying different estimators on the basis of the input set, for example, a minimal method if $n = m$ and least-squares fitting otherwise. Function $D: \Theta \times \mathcal{P} \to \mathbb{R}$ is the point-to-model residual function. Function $I : \mathcal{P}^* \times \Theta \times \mathbb{R} \to \mathcal{P}^*$ selects the inliers given model $\theta$ and threshold $\sigma$. For instance, if the original RANSAC approach is considered, 
$I_{\text{RANSAC}}(\theta,\sigma,\mathcal{P}) = \{ p \in \mathcal{P} \; | \; D(\theta, p) < \sigma \}$,  for truncated quadratic distance of MSAC, $I_{\text{MSAC}}(\theta,\sigma,\mathcal{P}) = \{ p \in \mathcal{P} \; | \; D^2(\theta, p) < 9 / 4 \sigma^2 \}$. The quality function is $Q : \mathcal{P}^* \times \Theta \times \mathbb{R} \to \mathbb{R}$. Higher quality is interpreted as better model.
For RANSAC, $Q_{\text{RANSAC}}(\theta,\sigma,\mathcal{P}) = |I(\theta,\sigma,\mathcal{P})|$ and for MSAC, it is 
\begin{equation*}
    Q_{\text{MSAC}}(\theta,\sigma,\mathcal{P}) = 
      \sum_{i = 1}^{|I(\theta,\sigma,\mathcal{P})|} \left(1 - \frac{D^2(\theta, I_i(\theta,\sigma,\mathcal{P}))}{9 / 4 \sigma^2} \right),
\end{equation*}
where $I_i(\theta,\sigma,\mathcal{P})$ is the $i$th inlier. 

\begin{table}
	\center
	\resizebox{0.999\columnwidth}{!}{\begin{tabular}{ c  l | c  l }
    \hline
   		\multicolumn{4}{ c }{ \cellcolor{black!10}Notation\rule{0pt}{2.0ex} } \\[0.2mm]
    \hline 
   		$\mathcal{P}$ & - Set of data points & 
   		$\sigma$ & - Noise standard deviation \\ 
   		$\theta$ & - Model parameters &
   		$D$ & - Residual function \\ 
   		$I$ & - Inlier selector function \; & 
   		$Q$ & - Model quality function \\ 
   		$F$ & - Fitting function &  
   		\; $m$ & - Minimal sample size \\
   		$\tau(\sigma)$ & - Inlier-outlier threshold &  
   		\; $\sigma_{\text{max}}$ & - Upper bound of $\sigma$ \\
    \hline     
\end{tabular}}
\label{tab:notation}
\end{table}

\section{Marginalizing sample consensus}

A method called MAGSAC is proposed in this section eliminating the threshold parameter from RANSAC-like robust model estimation. 


\subsection{Marginalization over $\sigma$}

Let us assume the noise $\sigma$ to be a random variable with density function $f(\sigma)$ and let us define a new quality function for model $\theta$ marginalizing over $\sigma$ as follows:
\begin{equation}
 Q^{*}(\theta,\mathcal{P})=\int Q(\theta,\sigma,\mathcal{P}) f(\sigma){\mathrm d \sigma}.
\end{equation}
Having no prior information, we assume $\sigma$ being uniformly distributed, $\sigma \sim  \mathcal{U}(0,\sigma_{max})$. Thus
\begin{equation}
 Q^{*}(\theta,\mathcal{P})=\frac{1}{\sigma_{max}}\int_0^{\sigma_{max}} Q(\theta,\sigma,\mathcal{P}) {\mathrm d \sigma}.
\end{equation}
For instance, using $Q(\theta,\sigma,\mathcal{P})$ of plain RANSAC, i.e.\ the number of inliers, where $\sigma$ is the inlier-outlier threshold and $\{D(\theta, p_{i})\}_{i=1}^{|\mathcal{P}|}$ are the distances to model $\theta$ such that 
$0 \leq D(\theta, p_{1})<D(\theta, p_{2})<....<D(\theta, p_{K})<\sigma_{max}<D(\theta, p_{K+1})<...<D(\theta, p_{|\mathcal{P}|})$  
we get a quality function
\begin{equation*}
    Q^{*}(\theta,\mathcal{P})= K- \frac{1}{\sigma_{max}}\sum_{k=1}^{K}D(\theta, p_{k})
    = \sum_{k=1}^{K} \left( 1 - \frac{D(\theta, p_{k})}{\sigma_{max}} \right).
\end{equation*}
 
Assuming the distribution of inliers and outliers to be uniform 
(inlier $\sim \mathcal{U}(0,\sigma)$; 
outlier $\sim \mathcal{U}(0,l)$)
and using log-likelihood of model $\theta$ as its quality function $Q$, we get
\begin{equation} 
    \begin{aligned}
        Q^{*}(\theta,\mathcal{P})= K(\ln {\frac{l}{\sigma_{max}}}+1)\\ -\frac{1}{\sigma_{max}}
        \sum_{k=1}^{K}D(\theta, p_{k}) (1+\ln {\frac{l}{D(\theta, p_{k})}})-|\mathcal{P}|\ln l.
    \end{aligned}
\end{equation}

Typically, the residuals of the inliers are calculated as the Eucledian-distance from the model in some $\rho$-dimensional space. In case of assuming errors of the distances along each axis of this $\rho$-dimensional space to be independent and normally distributed with the same variance $\sigma^2$,  $($residuals$)^2/ \sigma^2$ have chi-squared distribution with $\rho$ degrees of freedom.
Therefore, 
\begin{equation*}
    g(r \; | \; \sigma) = 2C(\rho)\sigma^{-\rho}\exp{(-r^2/2\sigma^2)}r^{\rho - 1}
\end{equation*}
is a density of residuals of inliers with
\begin{equation*}
    C(\rho) = \frac{1}{2^{\rho / 2}\Gamma(\rho / 2)},
\end{equation*}
where
\begin{equation*}
    \Gamma (a)=\int_{0}^{+\infty} t^{a-1}  \exp {(-t)} {\mathrm d } t
\end{equation*} 
for $a > 0$ is the gamma function.

In MAGSAC, the residuals of the inliers are described by a distribution with density $g(r \; | \; \sigma)$, and the outliers by a uniform one on the interval $[0,l]$. Note that, for images, $l$ can be set to the image diagonal.
The inlier-outlier threshold $\tau (\sigma)$ is set to the 0.95 or 0.99 quantile of the distribution with density $g(r \; | \; \sigma)$.
Consequently, the likelihood of model $\theta$ given $\sigma$ is 
\begin{align} 
    \normalsize
    \begin{aligned}
	    \text{L}(\theta,\mathcal{P} \; | \; \sigma) = \frac{1}{l^{|\mathcal{P}| - |\mathcal{I(\sigma)}|}} \\ \\
	    \prod_{p \in \mathcal{I(\sigma)}} \left[2 C(\rho) \sigma^{-\rho} D^{\rho - 1}(\theta, p) \exp \left(\frac{-D^2(\theta, p)}{2 \sigma^2} \right) \right].
    \end{aligned}
    \label{eq:posterior_density_with_product}
\end{align}
MAGSAC, for a given $\sigma$, uses log-likelihood of model $\theta$ as its quality function as follows: 
$Q(\theta,\sigma,\mathcal{P})= \ln { \text{L}(\theta,\mathcal{P} |\sigma) }$. 
Thus, the quality marginalized over $\sigma$ is the following.
\begin{equation} 
    \begin{aligned}
        Q_{\text{MAGSAC}}^{*}(\theta,\mathcal{P}) = \frac{1}{\sigma_{max}} \int_0 ^{\sigma_{max}}
        \ln { \text{L}(\theta,\mathcal{P} |\sigma) d\sigma} \\
\approx  -|\mathcal{P}| \ln{l} +\frac{1}{\sigma_{max}} \sum_{i=1}^{K} [  i(\ln {2C(\rho)l}-\rho\ln {\sigma_i})  \\
 -\frac{R_{i}}{\sigma_i^2} 
 +(\rho-1)Lr_{i} ](\sigma_i-\sigma_{i-1}),
    \end{aligned}
\end{equation}
where
$\{D(\theta, p_{i})\}_{i=1}^{|\mathcal{P}|}$ are the distances to model $\theta$, $\sigma_0=0$ and
$0\leq D(\theta, p_{1}) = \tau(\sigma_1) < D(\theta, p_{2}) = \tau(\sigma_2) < ... < D(\theta, p_{K}) = \tau(\sigma_K) < \tau(\sigma_{\text{max}}) < D(\theta, p_{K+1}) < ... < D(\theta, p_{|\mathcal{P}|})$, 
$R_i = \frac{1}{2}\sum_{j=1}^{i} D(\theta, p_{j})^2 $ and $Lr_i=\sum_{j=1}^{i} \ln D(\theta, p_{j})$.
As a consequence, the proposed new quality function $Q_{\text{MAGSAC}}^{*}$ does not depend on a manually set noise level $\sigma$. 

 
\subsection{$\sigma$-consensus model fitting}

Due to not having a set of inliers which could be used to polish the model obtained from a minimal sample, we propose to use weighted least-squares fitting where the weights are the point probabilities of being inliers.   

Suppose that we are given model $\theta$ estimated from a minimal sample. 
Let $\theta_{\sigma}=F(I(\theta,\sigma,\mathcal{P}))$ be the model implied by the inlier set $I(\theta,\sigma,\mathcal{P})$ selected using $\tau(\sigma)$ around the input model $\theta$. 
It can be seen from Eq.~\ref{eq:posterior_density_with_product} that the likelihood of point $p \in \mathcal{P}$ being inlier given model $\theta_\sigma$ is
\begin{equation*}
    \text{L}(p \; | \; \theta_\sigma, \sigma) = 2 C(\rho) \sigma^{-\rho} D^{\rho - 1}(\theta_\sigma, p) \exp \left(\frac{-D^2(\theta_\sigma, p)}{2 \sigma^2} \right).
\end{equation*}
For finding the likelihood of a point being an inlier marginalized over $\sigma$, the same approach is used as before:
\begin{equation}
    \begin{aligned}
        \text{L}(p \; | \; \theta) \approx \frac{2 C(\rho)}{\sigma_{max}}  
        \sum_{i = 1}^{K} (\sigma_i - \sigma_{i - 1}) \\ 
        \sigma_i^{-\rho} D^{\rho - 1}(\theta_{\sigma_i}, p) \exp \left(\frac{-D^2(\theta_{\sigma_i}, p)}{2 \sigma_i^2} \right).
    \end{aligned}
    \label{eq:point_weight}
\end{equation}
and the polished model $\theta_{\text{MAGSAC}}^*$ is estimated using weighted least-squares, where the weight of point $p \in \mathcal{P}$ is $\text{L}(p \; | \; \theta)$. 
 
\subsection{Termination criterion}

Not having an inlier set and, thus, at least a rough estimate of the inlier ratio, makes the standard termination criterion of RANSAC~\cite{hartley2003multiple} inapplicable, which is as follows: 
\begin{equation}
    k(\theta,\sigma,\mathcal{P}) = \frac{\ln (1 - \eta)}{\ln \left(1 - \left(\frac{|I(\theta,\sigma,\mathcal{P})|}{|\mathcal{P}|} \right)^{m} \right)},
\end{equation}
where $k$ is the iteration number, $\eta$ a manually set confidence in the results, $m$ the size of the minimal sample needed for the estimation, and $|I(\theta,\sigma,\mathcal{P})|$ is the inlier number of the so-far-the-best model. 

In order to determine $k$ without using a particular $\sigma$, it is a straightforward choice to marginalize similarly to the model quality. It is as follows:
\begin{align}
    \begin{split}
        k^*(\mathcal{P}, \theta) = \frac{1}{\sigma_{max}} \int_0 ^{\sigma_{max}}k(\theta,\sigma,\mathcal{P}) d\sigma  \\
        \approx \frac{1}{\sigma_{max}} \sum_{i = 1}^{K} \frac{(\sigma_i - \sigma_{i - 1}) \ln (1 - \eta)}{\ln \left(1 - \left(\frac{|I(\theta,\sigma_i,\mathcal{P})|}{|\mathcal{P}|} \right)^{m} \right)}.
    \end{split}
    \label{eq:iteration_number}
\end{align} 
Thus the number of iterations required for MAGSAC is calculated during the process and updated whenever a new so-far-the-best model is found, similarly as in RANSAC.

\section{Algorithms using $\sigma$-consensus}

\begin{figure}
  \centering
  \begin{subfigure}[t]{0.90\columnwidth}
 	 	\centering
  		\includegraphics[width=0.99\columnwidth]{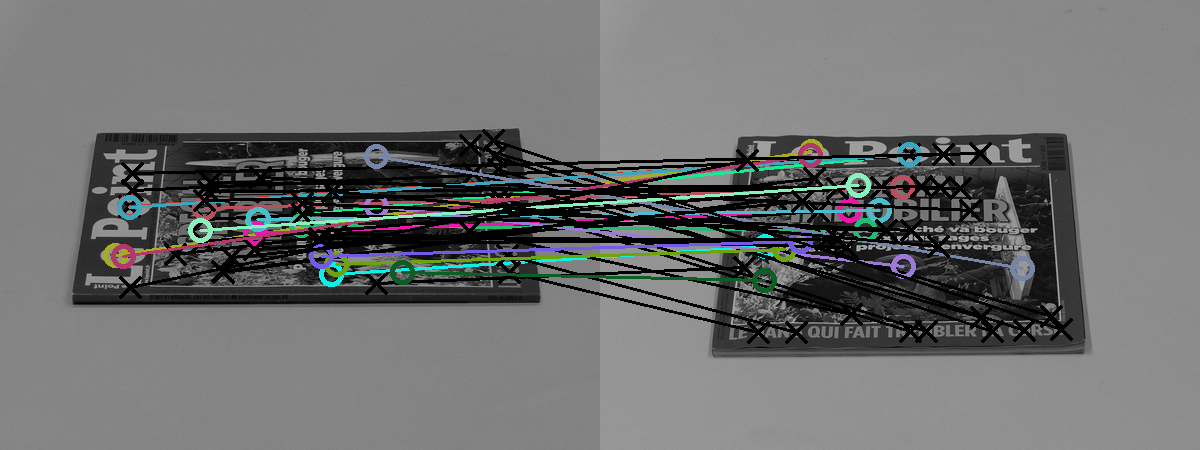}
        \caption{Homography; {\fontfamily{cmtt}\selectfont homogr} dataset. Errors: $\epsilon_{\text{LO-MSC}} = 4.3$ (2nd) and $\epsilon_{\text{MAGSAC}} = 2.9$ pixels (1st).}
  \end{subfigure}
  \begin{subfigure}[t]{0.90\columnwidth}
 	 	\centering
  		\includegraphics[width=0.99\columnwidth]{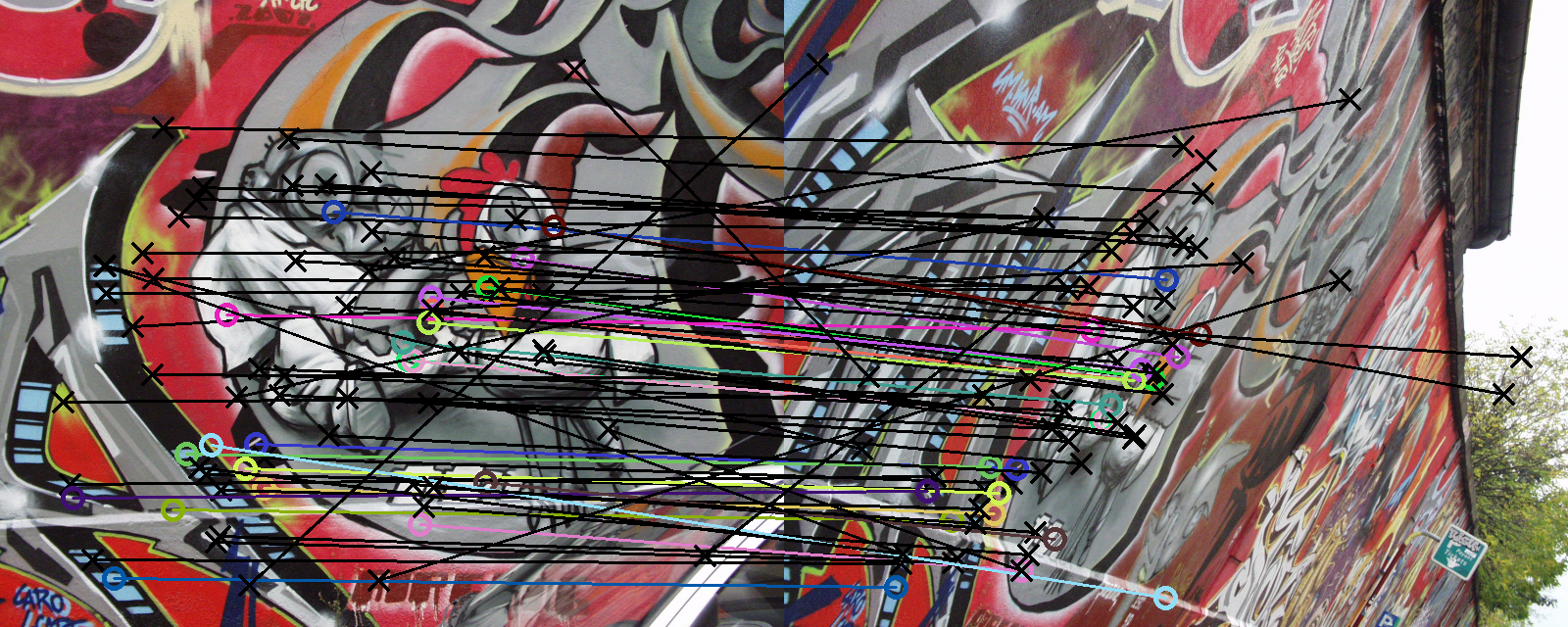}
        \caption{Homography; {\fontfamily{cmtt}\selectfont EVD} dataset. Errors: $\epsilon_{\text{LO-RSC}} = 9.1$ (2nd) and $\epsilon_{\text{MAGSAC}} = 4.4$ pixels (1st).}
  \end{subfigure}
  \begin{subfigure}[t]{0.90\columnwidth}
 	 	\centering
  		\includegraphics[width=0.99\columnwidth]{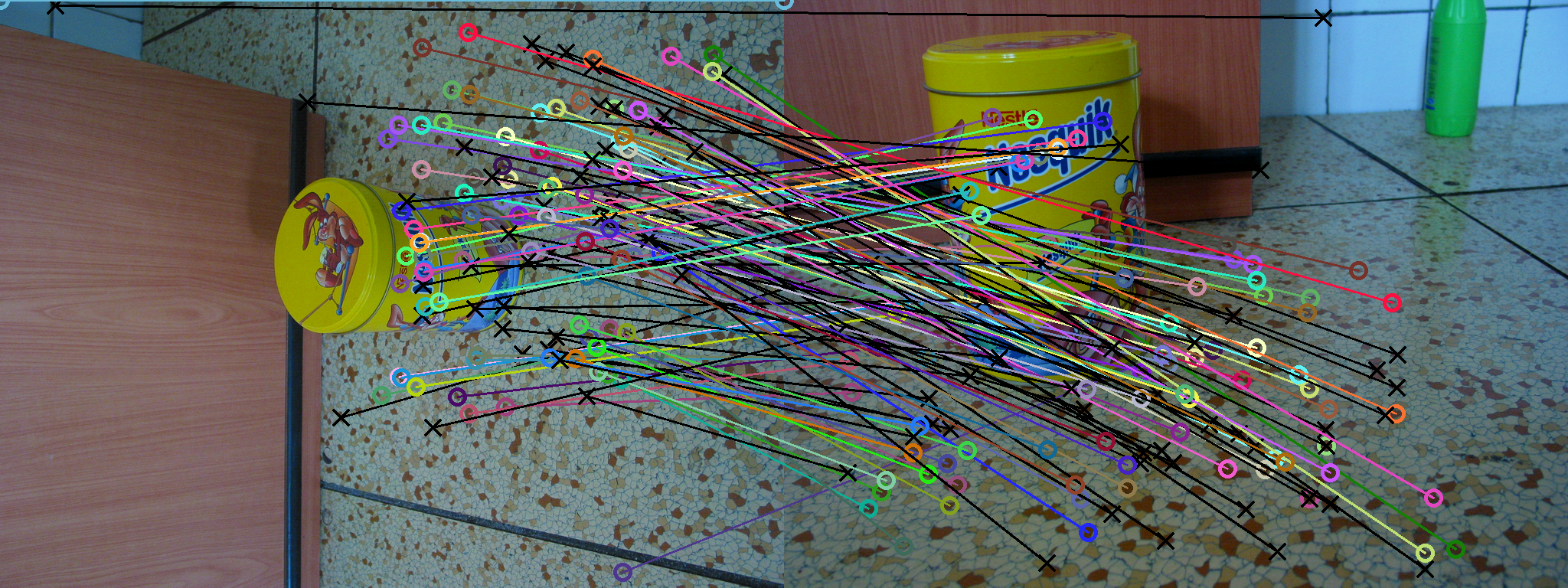}
        \caption{Fundamental matrix; {\fontfamily{cmtt}\selectfont kusvod2} dataset. Errors: $\epsilon_{\text{MSC}} = 14.3$ (2nd) and $\epsilon_{\text{MAGSAC}} = 0.5$ pixels (1st).}
  \end{subfigure}
  \begin{subfigure}[t]{0.90\columnwidth}
 	 	\centering
  		\includegraphics[width=0.99\columnwidth]{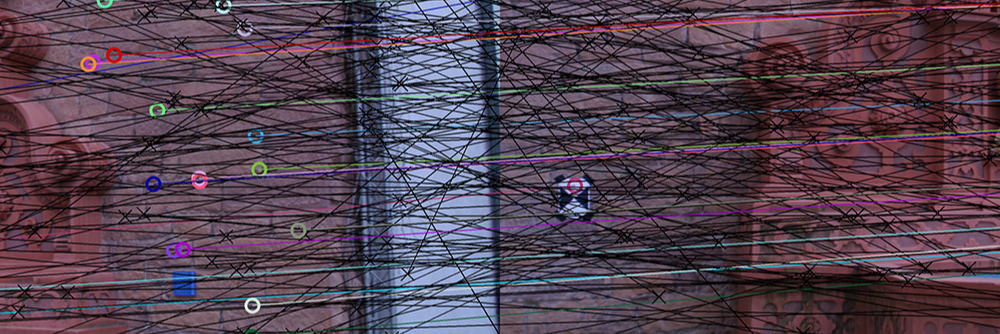}
        \caption{Essential matrix; {\fontfamily{cmtt}\selectfont Strecha} dataset. Errors: $\epsilon_{\text{MSC}} = 4.2$ (2nd) and $\epsilon_{\text{MAGSAC}} = 2.5$ pixels (1st).}
  \end{subfigure}
  \begin{subfigure}[t]{0.90\columnwidth}
 	 	\centering
  		\includegraphics[width=0.99\columnwidth]{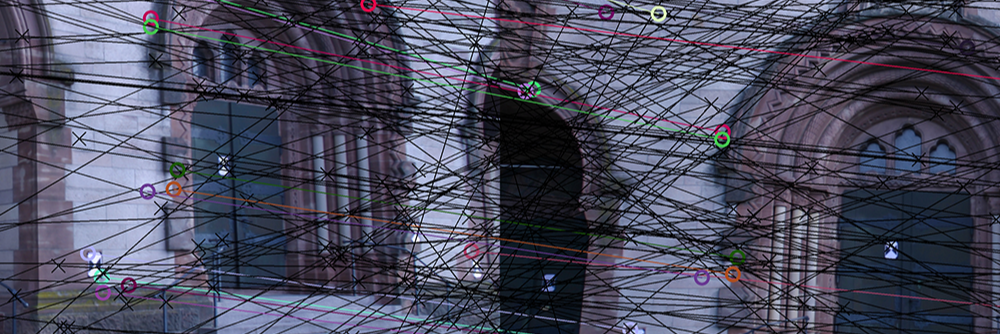}
        \caption{Essential matrix; {\fontfamily{cmtt}\selectfont Strecha} dataset. Errors: $\epsilon_{\text{MSC}} = 5.6$ (2nd) and $\epsilon_{\text{MAGSAC}} = 3.9$ pixels (1st).}
  \end{subfigure}
  \caption{ Example results of MAGSAC where it was significantly more accurate than the second most accurate method. Average errors (in pixels) are written in the captions. Inlier correspondences are drawn by color and outliers by black crosses.}
  \label{fig:example_images}
\end{figure}

In this section, we propose two algorithms applying $\sigma$-consensus.
First, MAGSAC will be discussed incorporating the proposed marginalizing approach, weighted least-squares and termination criterion. 
Second, a post-processing step is proposed which is applicable to the output of every robust estimator. In the experiments, it \textit{always improved} the input model without noticeable deterioration in the processing time, adding maximum a few milliseconds. 

\subsection{Speeding up the procedure}

Since plain MAGSAC would apply least-squares fitting a number of times, the implied computational complexity would be fairly high. Therefore, we propose techniques for speeding up the procedure.
In order to avoid unnecessary operations, we introduce a $\sigma_{\text{max}}$ value and use only the $\sigma$s smaller than $\sigma_{\text{max}}$ in the optimization procedure. Thus, from $\sigma_1 < \sigma_2 < ... < \sigma_K < \sigma_{max} < \sigma_{K+1} < ... < \sigma_{n}$ only $\sigma_1, \sigma_2, ...$, and $\sigma_i$ are used. This $\sigma_{max}$ can be set to a fairly big value, for example, 10 pixels.
In the case when the results suggest that $\sigma_{\text{max}}$ is too low, e.g.\ if the density mode of the residuals is close to $\sigma_{\text{max}}$, the computation can be repeated with a higher value. 

Instead of calculating $\theta_{\sigma_i}$ for every $\sigma_i$, we divide the range of $\sigma$s uniformly into $d$ partitions. Thus the processed set of $\sigma$s are the following: $\sigma_{1} + (\sigma_{\text{max}} - \sigma_{1}) / d$, $\sigma_{1} + 2 (\sigma_{\text{max}} - \sigma_{1}) / d$, $...$, $\sigma_{1} + (d - 1) (\sigma_{\text{max}} - \sigma_{1}) / d$, $\sigma_{\text{max}}$. By this simplification, the number of least-squares fittings drops to $d$ from $K$, where $d \ll K$.
In the experiments, $d$ was set to $10$. 

Also, as it was proposed for USAC~\cite{raguram2013usac}, there are several ways of skipping early the evaluation of models which do not have the chance of being better than the previous so-far-the-best. For this purpose, we apply SPRT~\cite{chum2008optimal} with a $\tau_{\text{ref}}$ threshold. Threshold $\tau_{\text{ref}}$ is not used in the model evaluation or inlier selection steps, but is used merely to skip applying $\sigma$-consensus when it is unnecessary. 
In the experiments, $\tau_{\text{ref}}$ was set to $1$ pixel.

Finally, the parallel implementation of $\sigma$-consensus can be straightforwardly done on GPU or multiple CPUs evaluating each $\sigma$ on a different thread. In our C++ implementation, it runs on multiple CPU cores.

\subsection{The $\sigma$-consensus algorithm} 

The proposed $\sigma$-consensus is described in Alg.~\ref{alg:post_processing}. 
The input parameters are: the data points ($\mathcal{P}$), initial model parameters ($\theta$), a user-defined partition number ($d$), and a limit for $\sigma$ ($\sigma_{\text{max}}$).

As a first step, the algorithm takes the points which are closer to the initial model than $\tau(\sigma_{\max})$ (line 1). Function $\tau$ returns the threshold implied by the input $\sigma$ parameter. In case of $\chi^2(4)$ distribution, it is $\tau(\sigma) = 3.64 \sigma$.
Then the residuals of the inliers are sorted, therefore, in $\{ \sigma_i \}_{i = 1}^{|\mathcal{I}|}$, $\sigma_i < \sigma_j \Leftrightarrow i < j$. In $\mathcal{I}_{\text{ord}}$, the indices of the points are ordered reflecting to $\{ \sigma_i \}_{i = 1}^{|\mathcal{I}|}$, thus $\sigma_i = D(\theta, \mathcal{I}_{\text{ord}, i}) / 3.64$ (line 2).
In lines 3 and 4, the weights are initialized to zero, and $\sigma_{\text{max}}$ is set to $\max (\{ \sigma_i \}_{i = 1}^{|\mathcal{I}|})$. 
Then the current $\sigma$ range is calculated.
For instance, the first range to process is $[\sigma_{1}, \sigma_{1} + \delta_\sigma]$. 
Note that $\sigma_{1} = 0$ due to having at least $m$ points at zero distance from the model.
The cycle runs from the first to the last point and, since $\mathcal{I}_{\text{ord}}$ is ordered, each subsequent point is farther from the model than the previous ones. 
Until the end of the current range, i.e.\ partition, is not reached (line 7), it collects the points (line 8) one-by-one.
After exceeding the boundary of the current range, $\theta_\sigma$ is calculated using all the previously collected points (line 10). Then, for each point, the weight is updated by the implied probability (line 12).
Finally, the algorithm jumps to the next range (line 13). 
After the weights have been calculated for each point, weighted least-squares fitting is applied to obtain the marginalized model parameters (line 14).

\subsection{MAGSAC} 

The MAGSAC procedure polishing every estimated model by $\sigma$-consensus is shown in Alg.~\ref{alg:magsac}. 
First, it initializes the model quality to zero and the required iteration number to $\infty$ (line 1).
In each iteration, it selects a minimal sample (line 3), fits a model to the selected points (line 4) validates it (line 5) and applies $\sigma$-consensus to obtain the parameters marginalized over $\sigma$ (line 6). 
The validation step includes degeneracy testing and tests which stop the evaluation of the model if there is no chance of being better than the previous so-far-the-best, e.g.\ by SPRT test~\cite{chum2008optimal}. Note that, for SPRT, the validation step is also included into $\sigma$-consensus when the distances from the current model are calculated (line 1 in Alg.~\ref{alg:post_processing}).
Finally, the model quality is calculated (line 8), the so-far-the-best model and required iteration number are updated (line 10) if required (line 9).
\textbf{As a post-processing step} in time sensitive applications, $\sigma$-consensus is a possible option for polishing the RANSAC output instead of applying a least-squares fitting to the inliers. In this case, $\sigma$-consensus is applied only once, thus improving the results without noticeable deterioration in the processing time. 


\begin{algorithm}
\begin{algorithmic}[1]
	\Statex{\hspace{-1.0em}\textbf{Input:} $\mathcal{P}$ -- points; $\theta$ -- model parameters; $d$ -- partition number; $\sigma_{\text{max}}$ -- $\sigma$ limit; $\eta$ -- confidence}
    \Statex{\hspace{-1.0em}\textbf{Output:} $\theta^*$ -- optimal model parameters}
   	\State{$\mathcal{I} \leftarrow I(\mathcal{P}, \theta, \tau(\sigma_{\text{max}}))$}
   	\State{$\mathcal{I}_{\text{ord}}, \{ \sigma_i \}_{i = 1}^{|\mathcal{I}|} \leftarrow \text{sort}(\{ D(\theta, p) \}_{p \in  \mathcal{I}})$} 
   	\State{$\{ w_i \}_{i = 1}^{|\mathcal{I}|} \leftarrow \{ 0 \}_{i = 1}^{|\mathcal{I}|}$, $\sigma_{\text{max}} \leftarrow \text{max}(\{ \sigma_i \}_{i = 1}^{|\mathcal{I}|})$}
   	\State{$\delta_\sigma \leftarrow \sigma_{\text{max}} / d$, $\sigma_{\text{next}} \leftarrow \delta_\sigma$, $\mathcal{I}_{\text{tmp}} \leftarrow \emptyset$}
   	\For{i = $1 \to |\mathcal{I}_{\text{ord}}|$}
        \State{$p \leftarrow \mathcal{I}_{\text{ord},i}$, $d_p \leftarrow D(\theta, p)$}
        \If {$d_p \leq \tau(\sigma_{\text{next}})$}
   	        \State{$\mathcal{I}_{\text{tmp}} \leftarrow \mathcal{I}_{\text{tmp}} \cup \{ p \}$}
            \State{\textbf{continue}}
        \EndIf
        \State{$\theta_\sigma \leftarrow F(\mathcal{I}_{\text{tmp}})$}
   	    \For{i = $1 \to |\mathcal{I}|$}
   	        \State{$w_i \leftarrow w_i + W(\theta_\sigma, \mathcal{I}_i, \delta_\sigma) / \sigma_{\max}$} \Comment{Eq.~\ref{eq:point_weight}}
   	    \EndFor
   	     \State{$\mathcal{I}_{\text{tmp}} \leftarrow \mathcal{I}_{\text{tmp}} \cup \{ p \}$, $\sigma_{\text{next}} \leftarrow \sigma_{\text{next}} + \delta_\sigma$}
    \EndFor
   	\State{$\theta^* \leftarrow F(\mathcal{I}, \{ w_i \}_{i = 1}^{|\mathcal{I}|})$} \Comment{Weighted LSQ}
\end{algorithmic}
\caption{\bf $\sigma$-consensus.}
\label{alg:post_processing}
\end{algorithm}

\begin{algorithm}
\begin{algorithmic}[1]
	\Statex{\hspace{-1.0em}\textbf{Input:} $\mathcal{P}$ -- data points; $\sigma_{\text{max}}$ -- $\sigma$ limit; $\sigma_{\text{ref}}$ -- reference $\sigma$; $m$ -- sample size; $d$ -- partition number; $\eta$ -- confidence}
    \Statex{\hspace{-1.0em}\textbf{Output:} $\theta^*$ -- optimal model; $q^*$ -- model quality}
   	\State{$q^* \leftarrow 0$, $k \leftarrow \infty$}
    \For{i = $1 \to k$}
        \State{$\{ p_j \}_{j = 1}^m \leftarrow$ Sample($\mathcal{P}$)}
        \State{$\theta \leftarrow F(\{ p_j \}_{j = 1}^m)$}
        \If {$\neg$Validate($\theta$, $\sigma_{\text{ref}}$)}	
            \State{\textbf{continue}}
        \EndIf
        \State{$\theta' \leftarrow \sigma\text{-consensus}(\mathcal{P}, \theta, d, \sigma_{\max})$} \Comment Alg.~\ref{alg:post_processing}
        \State{$q' \leftarrow Q(\theta', \mathcal{P})$}
        \If {$q > q^*$}
        	\State{ $q^*, \theta^*, k \leftarrow q', \theta', \text{Iters}(q', |\mathcal{P}|, \eta)$} \Comment Eq.~\ref{eq:iteration_number}
        \EndIf
   	\EndFor
\end{algorithmic}
\caption{\bf MAGSAC}
\label{alg:magsac}
\end{algorithm}

\section{Experimental Results}


To evaluate the proposed post-processing step, we tested several approaches with and without this step. The compared algorithms are: RANSAC, MSAC, LO-RANSAC, LO-MSAC, LO-RANSAAC~\cite{RaisFMMBC17}, and a contrario RANSAC~\cite{moisan2012automatic} (AC-RANSAC). LO-RANSAAC is a method including model averaging into robust estimation. AC-RANSAC estimates the noise $\sigma$. 
The same random seed was used for all methods and they performed a final least-squares on the obtained inlier set. The difference between RANSAC -- MSAC and LO-RANSAC -- LO-MSAC is merely the quality function. Moreover, the methods with LO prefix run the local optimization step proposed by Chum et al.~\cite{chum2003locally} with an inner RANSAC applied to the inliers. 
The parameters used are as follows: $\sigma = 0.3$ was the inlier-outlier threshold used for the RANSAC loop (this value was proposed in~\cite{lebeda2012fixing} and also suited for us). The number of inner RANSAC iterations was $r = 20$. The required confidence $\eta$ was $0.95$. There was a minimum number of iterations required (set to $20$) before the first LO step applied and also before termination. The reported error values are the root mean square (RMS) errors. For $\sigma$-consensus, $\sigma_{\text{max}}$ was set to $10$ pixels for all problems. The partition of $\sigma$ range was set to $d = 10$. Therefore, the processed set of $\sigma$s were $\sigma_{\text{max}}  / d$, $2 \sigma_{\text{max}} / d$, $...$, $(d-1) \sigma_{\text{max}} / d$, and $\sigma_{\text{max}}$. 


\subsection{Synthesized Tests}

To test the proposed method in a fully controlled environment, two cameras were generated by their $3 \times 4$ projection matrices $\textbf{P}_1 = \textbf{K}_1 [\textbf{I}_{3 \times 3} \; | \; 0]$ and $\textbf{P}_2 = \textbf{K}_2 [\textbf{R}_2 \; | -\textbf{R}_2 \textbf{t}_2]$. 
Camera $\textbf{P}_1$ was located in the origin and its image plane was parallel to plane $\text{XY}$. The position of the second camera was at a random point inside a unit-sphere around the first one, thus $|\textbf{t}_2| \leq 1$. 
Its orientation was determined by three random rotations affecting around the principal directions as follows:
$\textbf{R}_2 = \textbf{R}_{\text{X},\alpha} \textbf{R}_{\text{Y},\beta} \textbf{R}_{\text{Z},\gamma}$,
where $\textbf{R}_{\text{X},\alpha}$, $\textbf{R}_{\text{Y},\beta}$ and $\textbf{R}_{\text{Z},\gamma}$ are 3D rotation matrices rotating around axes X, Y and Z, by $\alpha$, $\beta$ and $\gamma$ degrees, respectively ($\alpha, \beta, \gamma \in [0, \pi / 2])$. 
%
%
Both cameras had a common intrinsic camera matrix with focal length $f_x = f_y = 600$ and principal points $[300, 300]^\trans$.
A 3D plane was generated with random tangent directions and origin $[0, 0, 5]^\trans$. It was sampled at $n_i$ locations, thus generating $n_i$ 3D points at most one unit far from the plane origin. These points were projected into the cameras. All of the random parameters were selected using uniform distribution. Zero-mean Gaussian-noise with $\sigma$ standard deviation was added to the projected point coordinates. Finally, $n_o$ outliers, i.e.\ uniformly distributed random point correspondences, were added. In total, $200$ points were generated, therefore $n_i + n_o = 200$.

\begin{figure*}
  \centering
  \begin{subfigure}[t]{0.63\columnwidth}
 	 	\centering
  		\includegraphics[width=0.999\columnwidth]{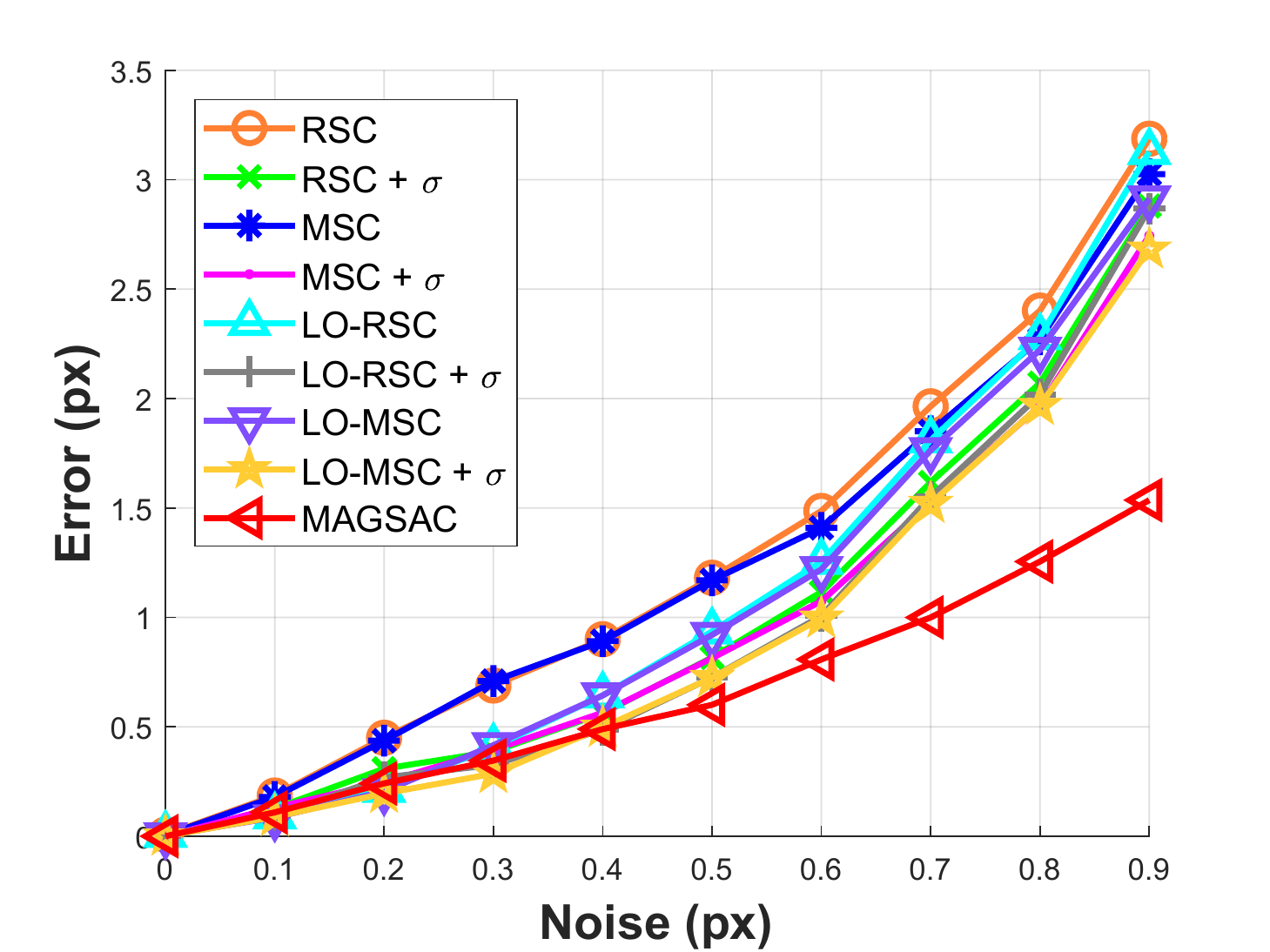}
        \caption{$20\%$ outl., $95\%$ conf.}
  \end{subfigure}\hfill
  \begin{subfigure}[t]{0.63\columnwidth}
 	 	\centering
  		\includegraphics[width=0.999\columnwidth]{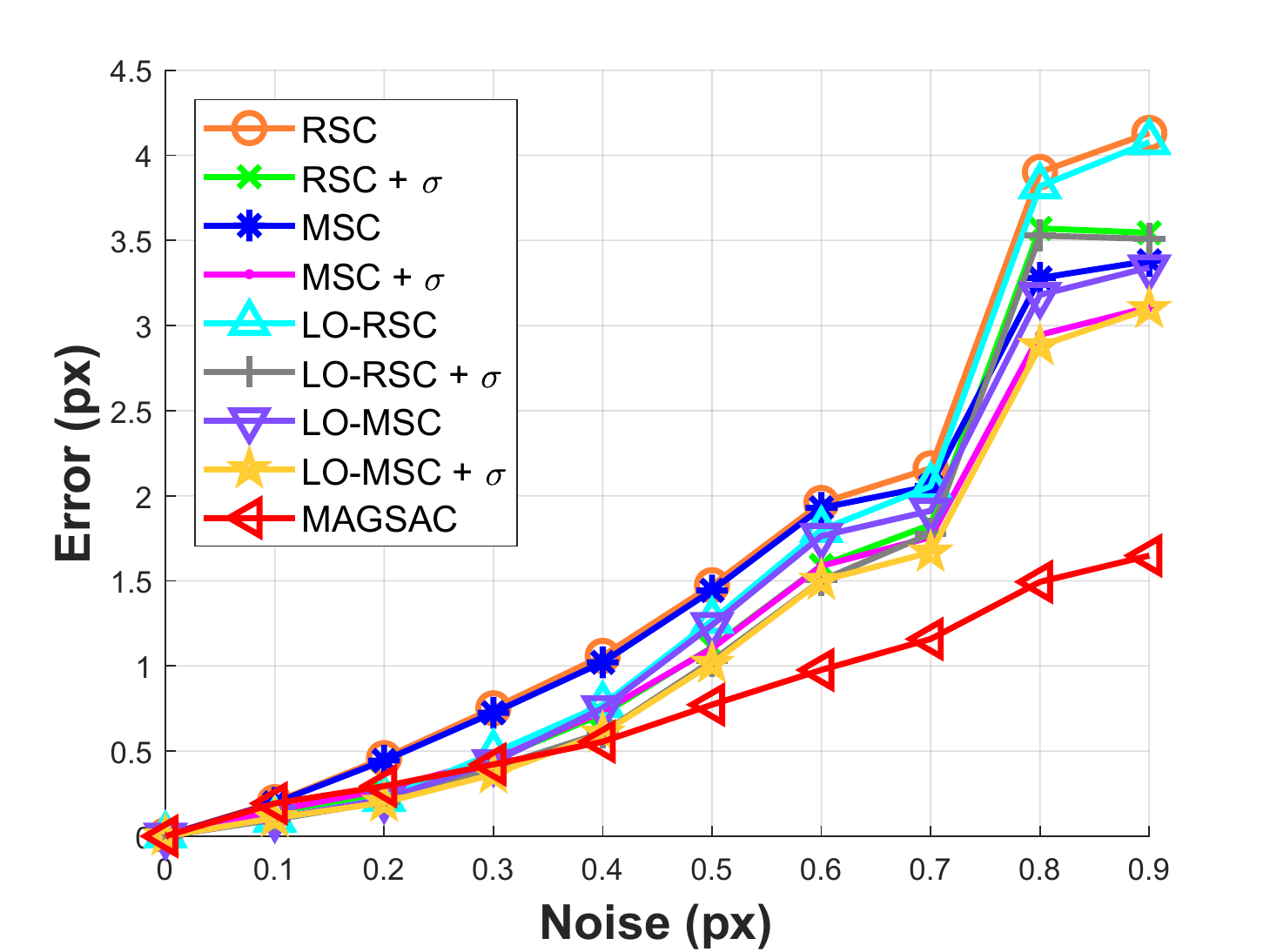}
        \caption{$50\%$ outl., $95\%$ conf.}
  \end{subfigure}\hfill
  \begin{subfigure}[t]{0.63\columnwidth}
 	 	\centering
  		\includegraphics[width=0.999\columnwidth]{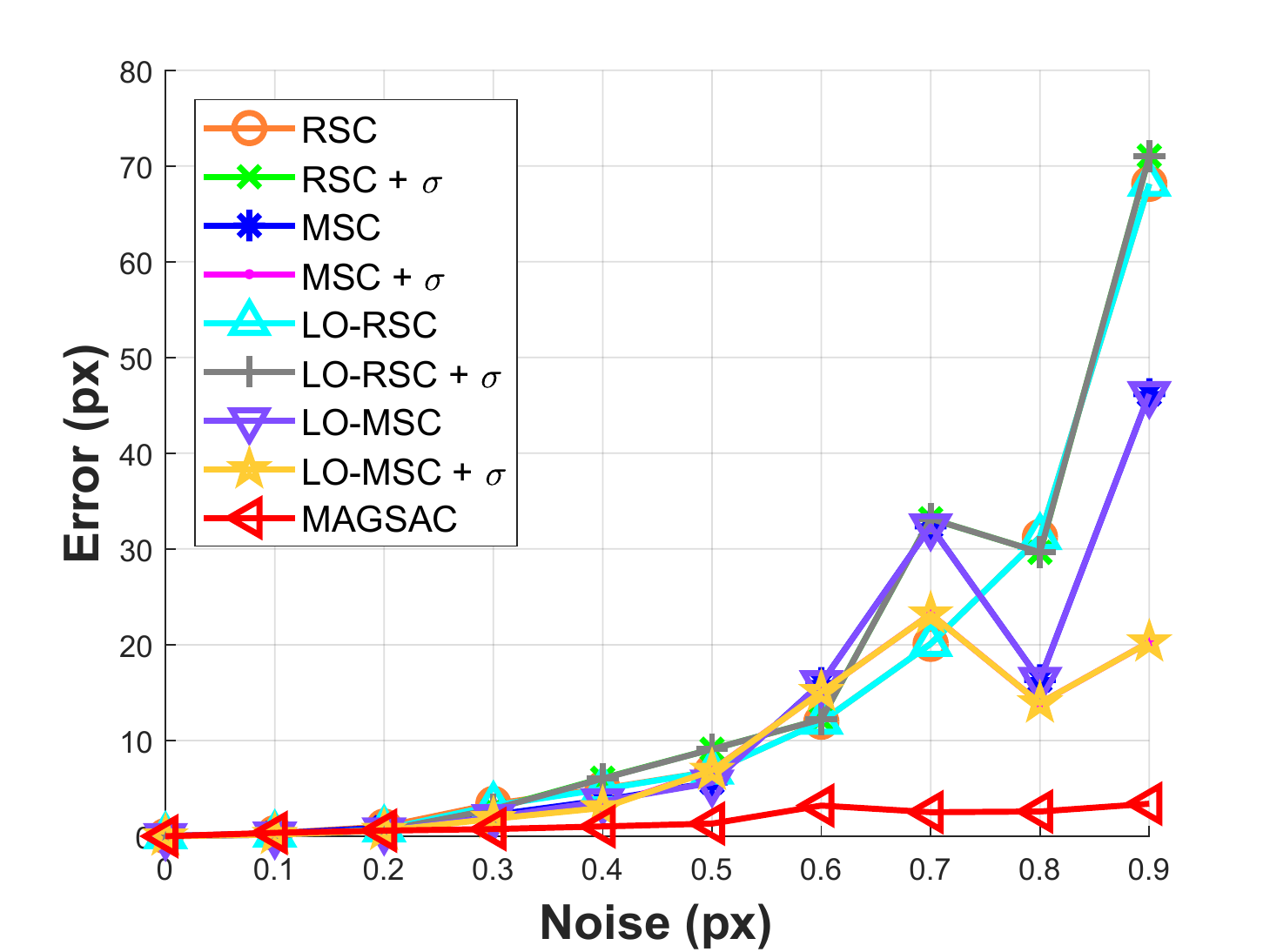}
        \caption{$80\%$ outl., $95\%$ conf.}
  \end{subfigure}\\ 
  \begin{subfigure}[t]{0.63\columnwidth}
 	 	\centering
  		\includegraphics[width=0.999\columnwidth]{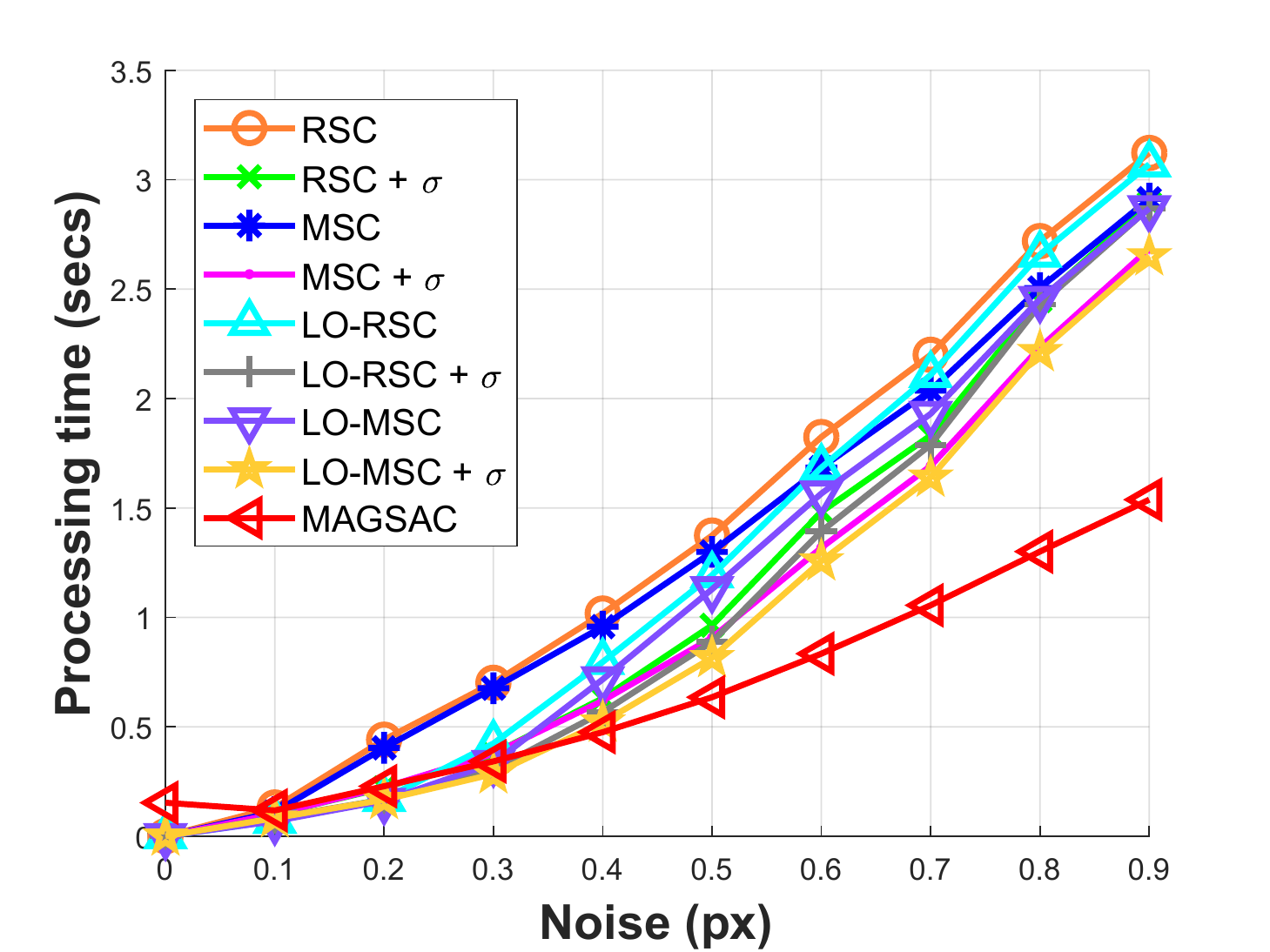} 
        \caption{$95\%$ conf.}
  \end{subfigure}\hfill
  \begin{subfigure}[t]{0.63\columnwidth}
 	 	\centering
  		\includegraphics[width=0.999\columnwidth]{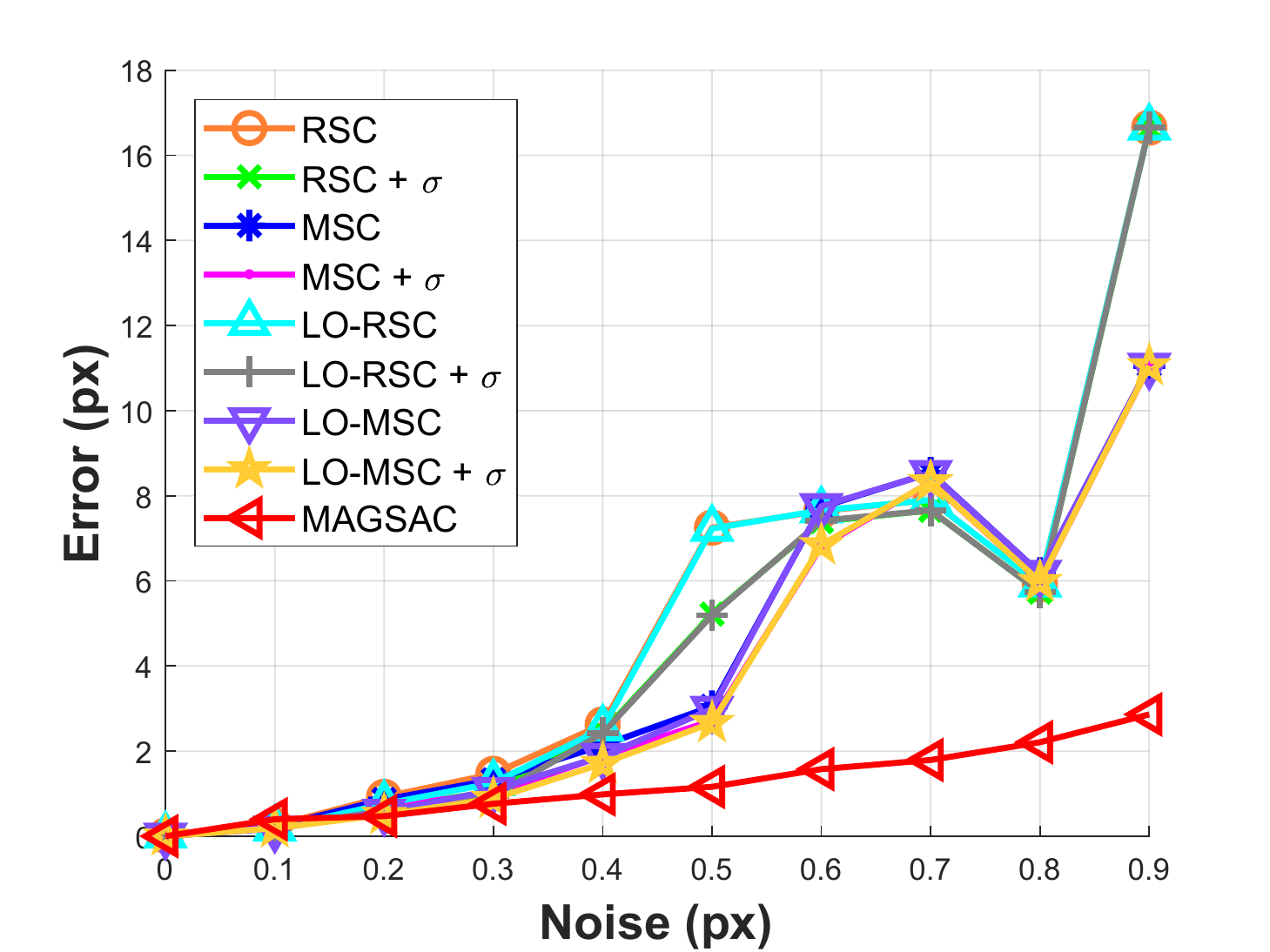}
        \caption{$80\%$ outl., 4\;314 iters.}
  \end{subfigure}\hfill
  \begin{subfigure}[t]{0.63\columnwidth}
 	 	\centering
  		\includegraphics[width=0.999\columnwidth]{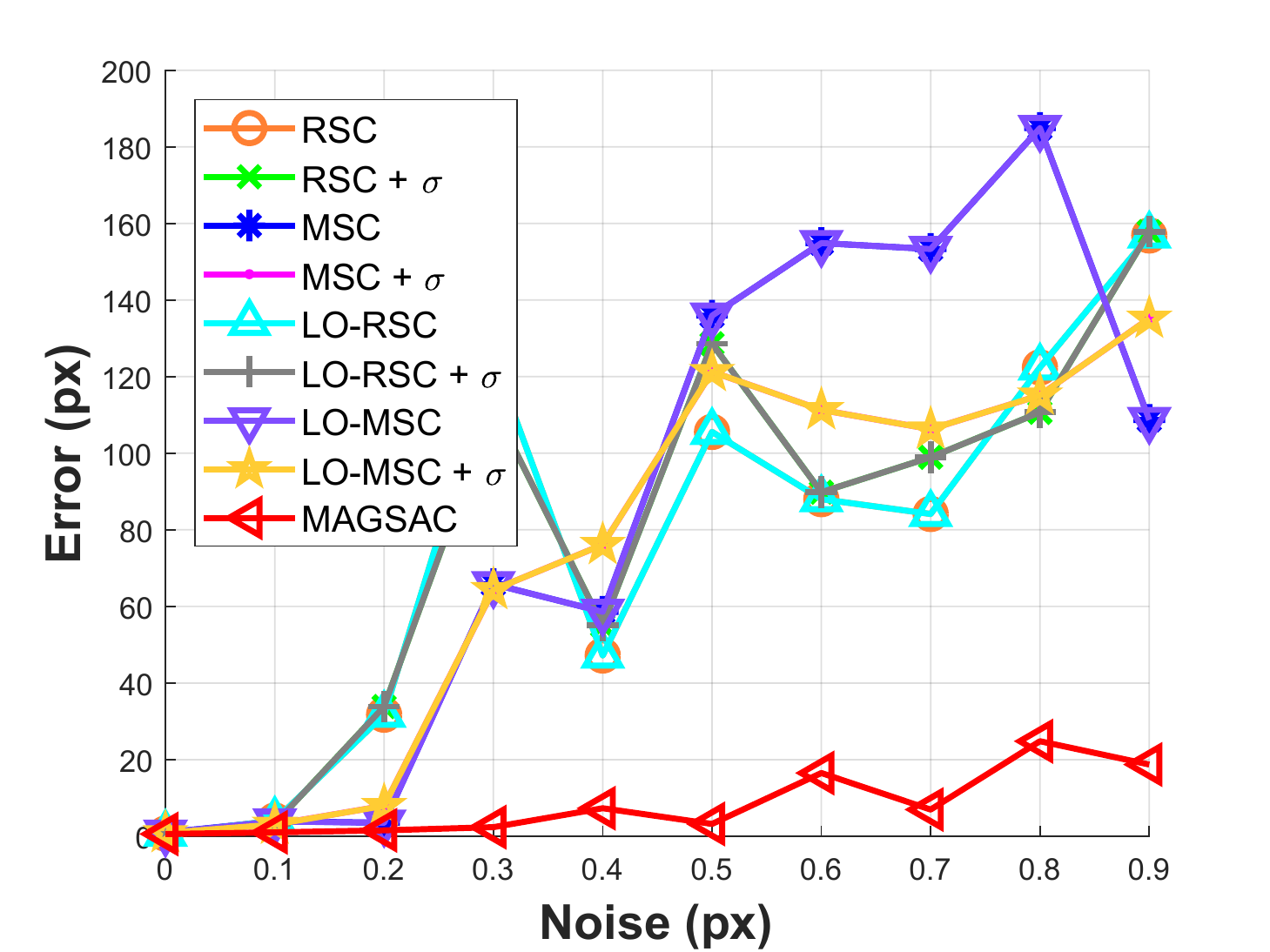}
        \caption{$90\%$ outl., 69\;074 iters.}
  \end{subfigure}
  \caption{ \textit{Synthetic homography fitting.} The competitor methods are: RANSAC, MSAC, LO-RANSAC, LO-MSAC and MAGSAC. Suffix "$+ \sigma$" means that $\sigma$-consensus was applied to the output. Plots (a--c) report the errors (in pixels) as function of the noise $\sigma$ with confidence set to $0.95$. Plot (d) shows the avg. processing time (in seconds). Plots (e--f) report the results made by using a fixed iteration number calculated from the ground truth inlier ratio and confidence set to  0.999.}
  \label{fig:synthetic_homography_fitting}
\end{figure*}

The mean results of $500$ runs are reported in Fig.~\ref{fig:synthetic_homography_fitting}. The competitor algorithms are: RANSAC (RSC), MSAC (MSC), LO-RANSAC (LO-RSC), LO-MSAC (LO-MSC) and MAGSAC. Suffix "$+ \sigma$" means that $\sigma$-consensus was applied as a post-processing step. 
Plots (a--c) reports the geometric accuracy (in pixels) as a function of the noise level $\sigma$ using different outlier ratios (a -- $0.2$, b -- $0.5$, c -- $0.8$). The RANSAC confidence was set to $0.95$.
For instance, outlier ratio $0.8$ means that $n_o = 160$ and $n_i = 40$. 
By looking at the differences between methods with and without the proposed post-processing step ("$+ \sigma$"), it can be seen that it almost always improved the results. 
E.g.\ the geometric error of LO-MSC is higher than that of LO-MSC + $\sigma$ for every noise $\sigma$.
MAGSAC results are superior to that of the competitor algorithms on every outlier ratio. 
It can be seen that it is less sensitive to noise and more robust to outliers. 
In (d), the processing time (in seconds) is reported as the function of the noise $\sigma$.
MAGSAC is the slowest on the easy scenes, i.e.\ when the noise $\sigma < 0.3$ pixels.
Thereafter, it becomes the fastest method due to requiring significantly fewer iterations than the others.
Plots (e--f) of Fig.~\ref{fig:synthetic_homography_fitting} demonstrate that the accuracy provided by MAGSAC cannot be achieved by simply letting RANSAC run longer. The charts report the results for a fixed iteration number, i.e.\ calculated from the ground truth inlier ratio and confidence set to 0.999. For outlier ratio 0.8, it was $\log(0.001) / \log(1 - 0.2^4) = 4\;314$. For outlier ratio 0.9, it was $\log(0.001) / \log(1 - 0.1^4) = 69\;074$.
It can be seen that MAGSAC obtains significantly more accurate results than the competitor algorithms. 
It finds the desired model in most of the cases even when the outlier ratio is high.

\subsection{Real World Experiments}

\begin{table*}
\center
  	\resizebox{0.99\linewidth}{!}{\begin{tabular}{| l | l | c || r | r || r | r || r | r || r | r || r | r |  r |  }
    \hline 
 	 	 \multicolumn{3}{|c||}{} & \multicolumn{1}{c|}{RSC} & \multicolumn{1}{c||}{+ $\sigma$} & \multicolumn{1}{c|}{MSC} & \multicolumn{1}{c||}{+ $\sigma$} & \multicolumn{1}{c|}{LO-RSC} & \multicolumn{1}{c||}{+ $\sigma$} & \multicolumn{1}{c|}{LO-MSC} & \multicolumn{1}{c||}{+ $\sigma$} & \multicolumn{1}{c|}{LO-RSAAC} & \multicolumn{1}{c|}{AC-RSC} & \multicolumn{1}{c|}{\textbf{MAGSAC}} \\ 
    \hline    
 	 	\multirow{4}{*}{\rot{{\fontfamily{cmtt}\selectfont \footnotesize kusvod2}}} & \multirow{4}{*}{\small \rot{$\mathbf{F}$, $24$}} & $e_{\text{avg}}$ & 0.73 & 0.60 & 0.75 & 0.64 & 0.56 & 0.52 & 0.58 & 0.50 & 1.01 & 0.63 & \textbf{0.38} \\
        & & $t$ & 38 & 39 & 19 & 19 & 25 & 25 & \textbf{17} & \textbf{17} & \textbf{17} & 55 & 31 \\
        & & $s$ & 661 & 661 & 313 & 313 & 316 & 316 & 160 & 160 & 160 & \textbf{71} &  382 \\
        & & \textit{fails} & 0.06 & 0.06 & 0.06 & 0.06 & 0.06 & 0.06 & 0.06 & 0.06 & 0.06 & 0.06 & \textbf{0.00} \\
    \hline    
 	 	\multirow{4}{*}{\rot{{\fontfamily{cmtt}\selectfont \footnotesize Adelaide}}} & \multirow{4}{*}{\small \rot{$\mathbf{F}$, $19$}} & $e_{\text{avg}}$ & 0.58 & 0.54 & 0.66 & 0.63 & 0.28 & \textbf{0.27} & 0.31 & 0.31 & 0.33 & 0.46 & 0.30 \\
        & & $t$ & 491 & 493 & 420 & 420 & 393 & 394 & \textbf{380} & \textbf{380} & \textbf{380} & 447 & 939 \\
        & & $s$ & 3\;327 & 3\;327 & 2\;752 & 2\;752 & 2\;221 & 2\;221 & 2\;091 & 2\;091 & 2\;091 & \textbf{2\;047} & 2\;638 \\
        & & \textit{fails} & \textbf{0.00} & \textbf{0.00} & \textbf{0.00} & \textbf{0.00} & \textbf{0.00} & \textbf{0.00} & \textbf{0.00} & \textbf{0.00} & \textbf{0.00} & \textbf{0.00} & \textbf{0.00} \\
    \hline   
 	 	\multirow{4}{*}{\rot{{\fontfamily{cmtt}\selectfont \footnotesize Multi-H}}} & \multirow{4}{*}{\rot{$\mathbf{F}$, $4$}} & $e_{\text{avg}}$ & 0.70 & 0.59 & 0.84 & 0.75 & 0.53 & 0.52 & 0.50 & 0.50 & 0.58 & 0.72 & \textbf{0.47} \\
        & & $t$ &  321 & 329 & 149 & 149 & 132 & 140 & 119 & 128 & 126 & \textbf{46} & 467 \\
        & & $s$ &  1\;987 & 1\;987 & 908 & 908 & 580 & 580 & 327 & 327 & 327 & \textbf{23} & 1\;324 \\
        & & \textit{fails} & \textbf{0.00} & \textbf{0.00} & \textbf{0.00} & \textbf{0.00} & \textbf{0.00} & \textbf{0.00} & \textbf{0.00} & \textbf{0.00} & \textbf{0.00} & \textbf{0.00} & \textbf{0.00} \\
    \hline    
 	 	\multirow{4}{*}{\rot{{\fontfamily{cmtt}\selectfont \footnotesize homogr}}} & \multirow{4}{*}{\small \rot{$\mathbf{H}$, $16$}} & $e_{\text{avg}}$ & 3.61 & 2.12 & 3.64 & 2.18 & 3.39 & 2.13 & 3.53 & 2.19 & 2.95 & 1.83 & \textbf{1.37} \\
        & & $t$ & 83 & 85 & \textbf{64} & 65 & 71 & 72 & 64 & 65 & 65 & 37 & 131 \\
        & & $s$ & 1\;815 & 1\;815 & 1\;395 & 1\;395 & 1\;478 & 1\;478 & 1\;222 & 1\;222 & 1\;222 & \textbf{148} & 877 \\
        & & \textit{fails} & 0.12 & 0.12 & 0.12 & 0.12 & 0.12 & 0.12 & 0.12 & 0.12 & 0.12 & \textbf{0.00} & 0.06 \\
    \hline    
 	 	\multirow{4}{*}{\rot{{\fontfamily{cmtt}\selectfont \footnotesize EVD}}} & \multirow{4}{*}{\small \rot{$\mathbf{H}$, $15$}} & $e_{\text{avg}}$ & 5.73 & 4.08 & 5.15 & 3.57 & 5.42 & 4.07 & 4.78 & 3.55 & 4.55 & 5.05 & \textbf{1.76} \\
 	 	 & & $t$ & 381 & 383 & 379 & 380 & 367 & 369 & 353 & 356 & 355 & 291 & \textbf{162} \\
 	 	 & & $s$ & 6\;212 & 6\;212 & 6\;106 & 6\;106 & 5\;847 & 5\;847 & 5\;540 & 5\;540 & 5\;540 & 3\;463 & \textbf{2\;239} \\
        & & \textit{fails} & 0.57 & 0.50 & 0.57 & 0.43 & 0.57 & 0.50 & 0.57 & 0.43 & 0.53 & 0.33 & \textbf{0.29} \\
    \hline 
 	 	\multirow{4}{*}{\rot{{\fontfamily{cmtt}\selectfont \footnotesize strecha}}} & \multirow{4}{*}{\small \rot{$\mathbf{E}$, $467$}} & $e_{\text{avg}}$ & 7.05 & 6.91 & 7.32 & 7.13 & 9.61 & 9.48 & 10.62 & 10.23 & 10.17 & 15.56 & \textbf{6.51} \\
 	 	 & & $t$ & 3\;046 & 3\;052 & 2\;894 & 2\;894 & 2\;548 & 2\;549 & 2\;535 & 2\;537 & 2\;536 & 4\;637 & \textbf{2\;398} \\
 	 	 & & $s$ & 3\;530 & 3\;530 & 3\;315 & 3\;315 & 2\;789 & 2\;789 & 2\;770 & 2\;770 & 2\;770 & 3\;680 & \textbf{2\;183} \\
        & & \textit{fails} & 0.24 & 0.22 & 0.26 & 0.22 & 0.27 & 0.22 & 0.26 & 0.22 & 0.24 & 0.23 & \textbf{0.00} \\
    \hline 
    \hline 
 	 	\multirow{4}{*}{\rot{{\fontfamily{cmtt}\selectfont \footnotesize all}}} &  & $e_{\text{avg}}$ &  3.07 & 2.47 & 3.04 & 2.48 & 3.30 & 2.83 & 3.39 & 2.88 & 3.27 & 4.04 & \textbf{1.80} \\
 	 	 & & $e_{\text{med}}$ & 2.17 & 1.36 & 2.24 & 1.47 & 1.98 & 1.33 & 2.06 & 1.35 & 1.98 & 1.28 & \textbf{0.92} \\
 	 	 & & $t$ & 727 & 730 & 654 & 655 & 589 & 592 & \textbf{578} & 581 & 580 & 921 & 688 \\
        & & \textit{fails} & 0.19 & 0.16 & 0.17 & 0.14 & 0.18 & 0.15 & 0.16 & 0.14 & 0.16 & 0.10 & \textbf{0.03} \\
    \hline 
\end{tabular}}
\caption{ \textit{Accuracy of robust estimators on two-view geometric estimation.} Fundamental matrix estimation ($\mathbf{F}$) on {\fontfamily{cmtt}\selectfont kusvod2} (24 pairs), {\fontfamily{cmtt}\selectfont AdelaideRMF} (19 pairs) and {\fontfamily{cmtt}\selectfont Multi-H} (4 pairs) datasets, homography estimation ($\mathbf{H}$) on {\fontfamily{cmtt}\selectfont homogr} (16 pairs) and {\fontfamily{cmtt}\selectfont EVD} (15 pairs) datasets, and essential matrix estimation ($\textbf{E}$) on the {\fontfamily{cmtt}\selectfont strecha} dataset (467 pairs). In total, the testing included $545$ image pairs.
The datasets, the problem, the number of the image pairs ($\#$) and the reported properties are shown in the first three columns. The other columns show the average results ($100$ runs on each image pair) of the competitor methods at $95\%$ confidence. Columns with "$+\sigma$" show the results when the proposed $\sigma$-consensus was applied to the output of the method on its left. The mean geometric error ($e_{\text{avg}}$; in pixels) of the estimated model w.r.t.\ the manually selected inliers are written in each 1st row; the mean processing time ($t$, in milliseconds) and the required number of samples ($s$) are written in every $2$nd and $3$rd rows. In the $4$th one, the proportion of failures, i.e.\ when the sough model is not found, is shown. The geometric error is the RMS Sampson distance for $\mathbf{F}$ and $\mathbf{E}$, and the RMS re-projection error for $\mathbf{H}$ using the ground truth inlier set. 
The thresholds proposed in~\cite{lebeda2012fixing} were used. For MAGSAC, $\sigma_{max} = 10$ pixels. }
\label{tab:dataset_comparison}
\end{table*}

In this section, MAGSAC and the proposed post-processing step is compared with state-of-the-art robust estimators on real-world data for fundamental matrix, homography and essential matrix fitting. See Fig.~\ref{fig:example_images} for example image pairs where the error ($\epsilon_{\text{MAGSAC}}$; in pixels) of the MAGSAC estimate was significantly lower than that of the second best method.  

\noindent \textbf{Fundamental Matrices.}  
To evaluate the performance on fundamental matrix estimation we downloaded {\fontfamily{cmtt}\selectfont kusvod2}\footnote{\url{http://cmp.felk.cvut.cz/data/geometry2view/}} (24 pairs), {\fontfamily{cmtt}\selectfont Multi-H}\footnote{\url{http://web.eee.sztaki.hu/~dbarath/}} (5 pairs), and  {\fontfamily{cmtt}\selectfont AdelaideRMF}\footnote{\url{cs.adelaide.edu.au/~hwong/doku.php?id=data}} (19 pairs) datasets. {\fontfamily{cmtt}\selectfont Kusvod2} consists of 24 image pairs of different sizes with point correspondences and fundamental matrices estimated from manually selected inliers. {\fontfamily{cmtt}\selectfont AdelaideRMF} and {\fontfamily{cmtt}\selectfont Multi-H} consist a total of 24 image pairs with point correspondences, each assigned manually to a homography or the outlier class. All points which are assigned to a homography were considered as inliers and the others as outliers. In total, $48$ image pairs were used from three publicly available datasets.  
All methods applied the seven-point method~\cite{hartley2003multiple} as a minimal solver for estimating $\textbf{F}$. Thus they drew minimal sets of size seven in each iteration. For the final least squares fitting, the normalized eight-point algorithm~\cite{hartley1997defense} was ran on the obtained inlier set. Note that all fundamental matrices were discarded for which the oriented epipolar constraint~\cite{chum2004epipolar} did not hold.

The first three blocks of Table~\ref{tab:dataset_comparison}, each consisting of three rows, report the quality of the estimation on each dataset as the average of 100 runs on every image pair. The first two columns show the name of the tests and the investigated properties: 
\textbf{(1)} $e_{\text{avg}}$ is the RMS geometric error in pixels of the obtained model w.r.t.\ the manually annotated inliers. For fundamental matrices and homographies, it is the average Sampson distance and re-projection error, respectively. For essential matrices, it is the mean Sampson distance of the implied $\textbf{F}$ and the correspondences. 
\textbf{(2)} Value $t$ is the mean processing time in milliseconds. 
\textbf{(3)} Value $s$ is the mean number of samples, i.e.\ RANSAC iterations, had to be drawn till termination. Note that the iteration numbers of methods applied with or without the proposed post-processing are equal.  

It can be seen that for $\textbf{F}$ estimation the proposed post-processing step improved the results in nearly all of the tests with negligible deterioration in the processing time. 
The errors were reduced by approximately $8\%$ compared with the methods without $\sigma$-consensus. MAGSAC led to the most accurate results for {\fontfamily{cmtt}\selectfont kusvod2} and {\fontfamily{cmtt}\selectfont Multi-H} datasets and it was the third best for {\fontfamily{cmtt}\selectfont AdelaideRMF} dataset by a small margin of $0.03$ pixels.

\noindent \textbf{Homographies.}  
To test homography estimation we downloaded {\fontfamily{cmtt}\selectfont homogr} (16 pairs) and {\fontfamily{cmtt}\selectfont EVD}\footnote{\url{http://cmp.felk.cvut.cz/wbs/}} (15 pairs) datasets. Each consists of image pairs of different sizes from $329 \times 278$ up to $1712 \times 1712$ with point correspondences and inliers selected manually.   
The {\fontfamily{cmtt}\selectfont Homogr} dataset consists of mostly short baseline stereo images, whilst the pairs of {\fontfamily{cmtt}\selectfont EVD} undergo an extreme view change, i.e.\ wide baseline or extreme zoom. 
All algorithms applied the normalized four-point algorithm~\cite{hartley2003multiple} for homography estimation both in the model generation and local optimization steps. 
%
The $4$th and $5$th blocks of Fig.~\ref{tab:dataset_comparison} show the mean results computed using all the image pairs of each dataset. Similarly as for $\textbf{F}$ estimation, the proposed post-processing step always improved (by $1.42$ pixels on average). 
For both datasets, the results obtained by MAGSAC were significantly more accurate than what the competitor algorithms obtained.

\noindent \textbf{Essential Matrices.}  
To estimate essential matrices, we used the {\fontfamily{cmtt}\selectfont strecha} dataset~\cite{strecha2004wide} consisting of image sequences of buildings. 
All images are of size $3072 \times 2048$. The ground truth projection matrices are provided. 
The methods were applied to all possible image pairs in each sequence. 
The SIFT detector~\cite{lowe1999object} was used to obtain correspondences. 
For each image pair, a reference point set with ground truth inliers was obtained by calculating $\textbf{F}$ from the projection matrices~\cite{hartley2003multiple}. Correspondences were considered as inliers if the symmetric epipolar distance was smaller than $1.0$ pixel. All image pairs with less than $50$ inliers found were discarded. In total, $467$ image pairs were used in the evaluation. 
The results are reported in the $6$th block of Table~\ref{tab:dataset_comparison}. The trend is similar to the previous cases. 
The most accurate essential matrices were obtained by MAGSAC. 
Also it was the fastest algorithm on average. 

%
%
%


\section{Conclusion}

A robust approach, called $\sigma$-consensus, was proposed for eliminating the need of a user-defined threshold by marginalizing over a range of noise scales. Also, due to not having a set of inliers, a new model quality function and termination criterion were proposed. 
Applying $\sigma$-consensus, we proposed two methods: first, MAGSAC applying $\sigma$-consensus to each of the models estimated from a minimal sample. The method is superior to the state-of-the-art in terms of geometric accuracy on publicly available real-world datasets for epipolar geometry (both $\textbf{F}$ and $\textbf{E}$) and homography estimation. 
The method is often faster than other RANSAC variants in case of high outlier ratio. 
The proposed post-processing step applies $\sigma$-consensus only once: to polish the RANSAC output. The method nearly \textit{always improved} the model quality on a wide range of vision problems without noticeable deterioration in processing time, i.e.\ at most a few milliseconds. We see no reason for not applying it after the robust estimation finished.

\section{Acknowledgement}

This work was supported by the OP VVV project CZ.02.1.01/0.0/0.0/16019/000076 Research Center for Informatics, by the Czech Science Foundation grant GA18-05360S, and by the Hungarian Scientic Research Fund (No. NKFIH OTKA KH-126513).

{\small
\bibliographystyle{ieee_fullname}
\bibliography{egbib}
}

\end{document}